\documentclass{article}
\usepackage{spconf,amsmath,epsfig,amsfonts,eufrak,amssymb,pifont,color,fancyhdr}
\usepackage[preprint]{spconfa4}

\copyrightnotice{}

\let\OLDthebibliography\thebibliography
\renewcommand\thebibliography[1]{
  \OLDthebibliography{#1}
  \setlength{\parskip}{0pt}
  \setlength{\itemsep}{0pt plus 0.3ex}
}
 
\begin{document}
\topmargin = 0mm
\sloppy

\def\x{{\mathbf x}}
\def\L{{\cal L}}

\title{Distortion-Tolerant Monocular Depth Estimation On Omnidirectional Images Using Dual-cubemap}
%
\name{Zhijie Shen$^{\dagger}$, Chunyu Lin$^{\dagger*}$, Lang Nie$^{\dagger}$, Kang Liao$^{\dagger}$ and Yao Zhao$^{\dagger}$}
\address{$^{\dagger}$Institute of Information Science, Beijing Jiaotong University, Beijing 100044, China}

\maketitle                       
 
\thispagestyle{fancy}            
\fancyhead{}                      
\lfoot{*Corresponding author.}                
\rfoot{978-1-6654-3864-3/21/\$31.00~\copyright 2021 IEEE}                 
\cfoot{\quad}                    
 
\renewcommand{\headrulewidth}{0pt} 
\renewcommand{\footrulewidth}{0pt}
 

%
\begin{abstract}
Estimating the depth of omnidirectional images is more challenging than that of normal field-of-view (NFoV) images because the varying distortion can significantly twist an object's shape. The existing methods suffer from troublesome distortion while estimating the depth of omnidirectional images, leading to inferior performance. To reduce the negative impact of the distortion influence, we propose a distortion-tolerant omnidirectional depth estimation algorithm using a dual-cubemap. It comprises two modules: Dual-Cubemap Depth Estimation (DCDE) module and Boundary Revision (BR) module. In DCDE module, we present a rotation-based dual-cubemap model to estimate the accurate NFoV depth, reducing the distortion at the cost of boundary discontinuity on omnidirectional depths. Then a boundary revision module is designed to smooth the discontinuous boundaries, which contributes to the precise and visually continuous omnidirectional depths.
Extensive experiments demonstrate the superiority of our method over other state-of-the-art solutions.
\end{abstract}
\section{Introduction}
\label{sec:intro}
Compared with NFoV images, omnidirectional images contain 360° FoV environmental information, representing a 3D scene as a 2D equirectangular map. Benefiting from the advantage of wide FoV, the 360° camera gains increasing applications in indoor robotic systems, autonomous driving, and virtual reality. Consequently, estimating the depth directly on an omnidirectional image is appealing and demanding for other subsequent computer vision tasks.

The existing monocular depth estimations have achieved encouraging improvements on NFoV images due to the advance of deep learning and availability of large-scale 3D training data. But these excellent methods cannot be directly transferred to estimate the depth of an omnidirectional image. Even FRCN ~\cite{2016Deeper}, the state-of-the-art monocular depth estimation method, exhibits poor performance in omnidirectional images because of the geometric distortion. In general, omnidirectional images are constructed by extending a sphere to a 360°$\times$180° equirectangular map, with increasing distortion from the equator to poles. The distortion twists the shapes and structures of objects to a varying degree and affects the depth estimation performance. 

  \begin{figure}[!t]
  \centering
  \includegraphics[width=.5\textwidth]{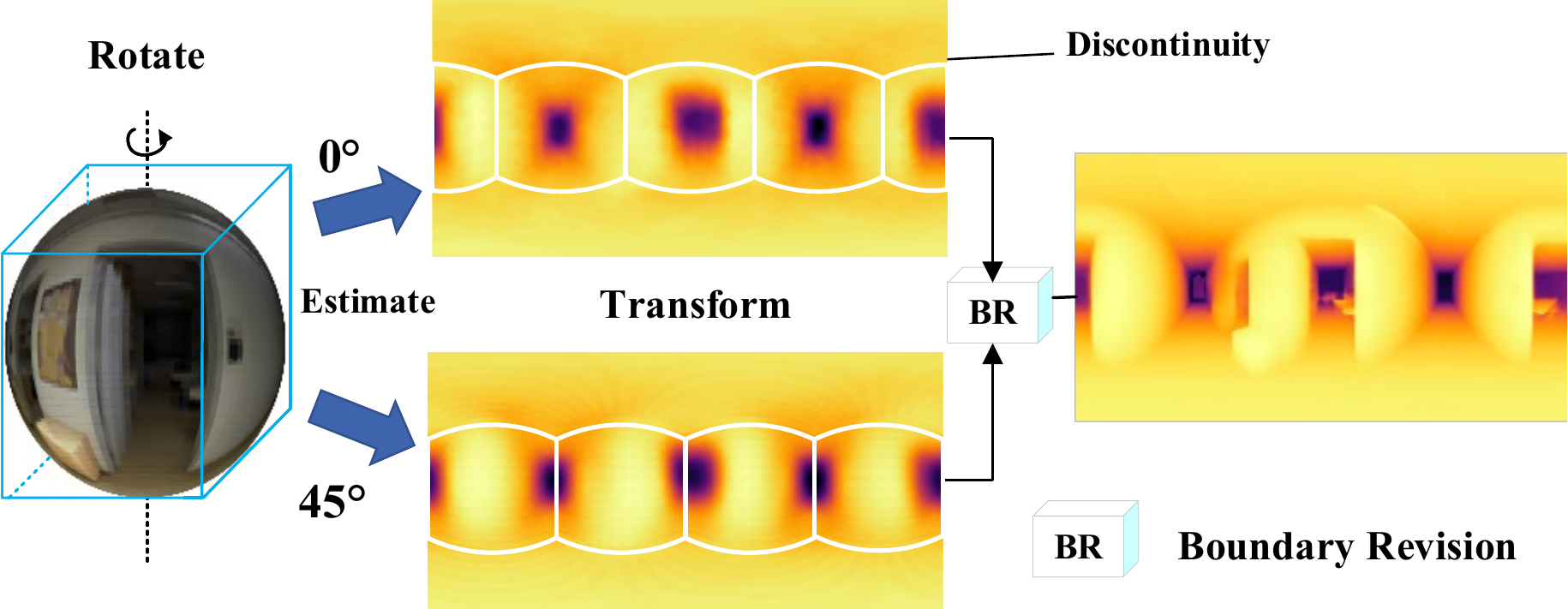} 
  \caption{An overview of our method.} 
  \label{fig_earth} 
\end{figure}
       The existing omnidirectional depth estimation methods have to suffer from the effects of distortion while estimating the depth. Some researches~\cite{Su2020Kernel,Esteves2020Learning,Cohen2018Spherical,Coors2018SphereNet} propose to estimate the omnidirectional depth directly from a single equirectangular projection map using convolutional neural networks (CNNs). With the powerful feature extraction capability of CNNs, the depth around the equator can be predicted accurately, but the performance drops sharply in the areas with large distortions due to the limited receptive of the network. Considering the regularity of distortion distribution in an omnidirectional image, Su $et\ al.$ ~\cite{Su2017Flat2Sphere} design convolutional kernels of different sizes to deal with the distortion from different regions. Bi-Projection Fuse (BiFuse) ~\cite{2020BiFuse} is proposed to estimate omnidirectional depth by combining an equirectangular map with a cubemap, further reducing the effects of distortion. 
       
       To eliminate the effects of distortions as much as possible, we propose a distortion-tolerant omnidirectional depth estimation method. In particular, we estimate the depth on the cubemap instead of the distorted equirectangular map. However, when the equirectangular map is converted into the cubemap, there is a 25$\%$ loss of pixels. In addition, the estimated depth of each face in this cubemap is not continuous. To tackle these two problems, we adopt another 45° rotating cubemap to reduce the pixel loss and smooth the depth difference simultaneously. Based on the dual-cubemap, our solution has more tolerance to distortions than the existing schemes.
       
       Our framework comprises two main modules: cube depth estimation and boundary revision. In the cube depth estimation module, we estimate the depth from dual-cubemap, where the two cubemaps are obtained by rotating 45° from the same sphere. Also, a boundary-aware block is designed for information interaction between the two branches of the dual-cubemap, focusing on the cubemap boundary and mutually promoting depth estimation from the other branch. Then, the depth of cubemap is converted back to the equirectangular format, and an encoder-decoder network revises the coarse omnidirectional depth in the boundary revision module.
       
       In experiments, we test our method on two real indoor scene omnidirectional datasets, Matterport3D~\cite{Chang2018Matterport3D} and Stanford2D3D~\cite{DBLP:journals/corr/ArmeniSZS17}. The results show that the proposed framework outperforms the current state-of-the-art methods. 
       
       Our contributions are summarized as follows:
\begin{itemize}
  \item We design an end-to-end framework consisting of a DCDE module and a BR module, aiming to eliminate the effects of distortions and revise the discontinuous boundaries, respectively.
  \item Different from the existing methods, our method estimates the omnidirectional depth from an interactive dual-cubemap, contributing to distortion-tolerant depth estimated results.
  \item  Experiments demonstrate that the proposed method exceeds the current state-of-the-art methods.
\end{itemize}
\begin{figure}[t]
  \centering
  \includegraphics[width=.35\textwidth]{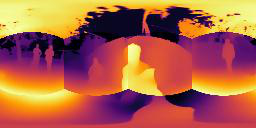} 
  \caption{An example of the discontinuity.} 
  \label{fig_cover} 
\end{figure}
\section{Related Work}
We briefly discuss the existing methods related to our work in this section.
\subsection{Monocular Depth Estimation}
Monocular depth estimation, a process of mapping from low to high dimensions, can be very challenging. Early researchers solve the mapping function in a hand-crafted features manner. Nowadays, many recent approaches extract features using deep learning techniques. Eigen $et\ al.$~\cite{DBLP:journals/corr/EigenPF14} propose the first CNNs based network. They adopted a coarse-to-fine strategy by designing a multi-scale module to improve predicted results. Inspired by ~\cite{2018Fully}, Laina $et\ al.$~\cite{Laina2016Deeper} merged pre-trained ResNet-50~\cite{2016Deep} and up-sampling blocks into FRCN~\cite{Laina2016Deeper} to estimate the depth. Considering the auxiliary role of other visual tasks, Jiao $et\ al.$~\cite{DBLP:conf/eccv/JiaoCSL18} introduced semantic information and designed an attention-driven loss as a bond. Different from treating depth estimation as a regression task, the approach of Fu $et\ al.$~\cite{2018Deep} adopted an ordinal regression strategy to transform depth estimation into a classification task by discretizing depth value. However, all the methods mentioned above are designed for NFoV images, and can not be directly applied to our task.
\subsection{Omnidirection Depth Estimation}

Solving distortion is the key to applying traditional CNNs based approaches. Previous researches attempted to make it by designing various distortion-aware structures. Noticing that the distortion changes from the equator to the poles, Su $et\ al.$~\cite{Su2017Flat2Sphere} designed convolution kernels of different sizes to deal with the distortion from different regions. However, the relationship between the convolution kernels' size and the distortion regions is not a simple linear mapping. Zioulis $et\ al.$~\cite{Zioulis2018OmniDepth} proposed OmniDepth to adopt the spherical layer in ~\cite{Su2017Flat2Sphere} as the pre-processing module. In the omnidirectional target detection task, Benjamin $et\ al.$~\cite{Coors2018SphereNet} proposed SphereNet to correct distortion by introducing distortion invariance and spherical sampling position mode. DACF~\cite{Tateno_2018_ECCV} designed a distortion-aware deformable convolution filter to directly perform depth regression on panoramas, getting a more accurate result. 

On the other hand, some estimate omnidirectional depth using cubemap. This representation suffers from less distortion but brings discontinuity on boundaries for each face. To overcome this issue, Cheng $et\ al.$~\cite{Cheng2018Cube} propose cube padding (cp) to utilize the connectivity between faces of the cube for image padding. Nevertheless, they ignore the characteristic of perspective projection. Spherical padding (sp) is presented in ~\cite{2020BiFuse} to reduce the boundary discontinuity and combined with an equirectangular map to estimate depth. Unfortunately, the equirectangular format introduces distortion again, which inevitably leads to the instability of their training. 
\section{Our method}
This section describes the proposed method in detail, and the architecture is illustrated in Fig. 3. Firstly, we introduce our motivation in Section 3.1. The proposed dual-cube depth estimation module and the boundary revision module are discussed in Section 3.2 and Section 3.3. Finally, we formulate the loss functions in Section 3.4.

  \begin{figure*}[!ht]
 \label{fig_pipeline}
 \centering
 \includegraphics[width=\textwidth]{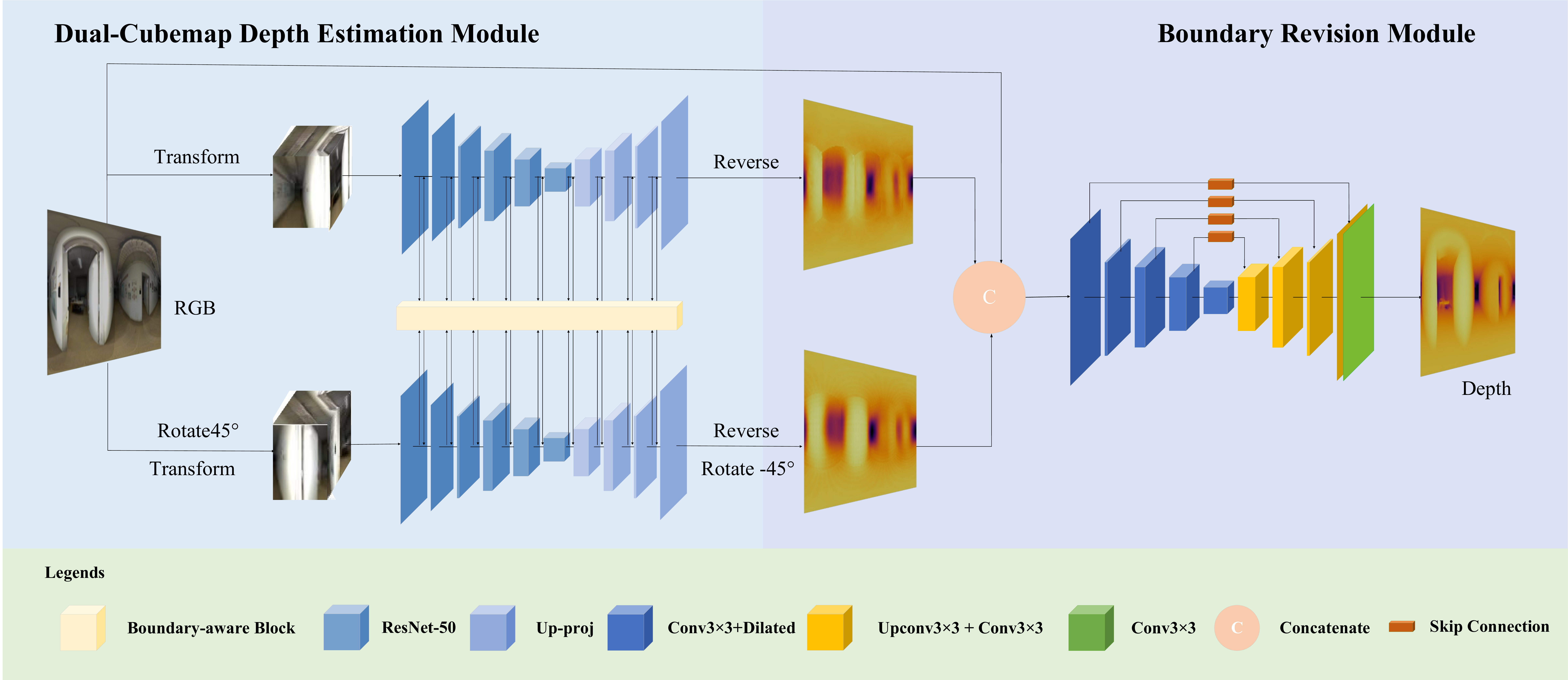}\\
 \caption{An overview of our framework: 1) Input image is used to another equirectangular representation by a rotation of 45°. 2) Two maps are converted into cubemap representation and sent to the DCDE module. 3) The output of DCDE is adjusted to be internally consistent, concatenated and sent to the BR module for improvement.
 }
\end{figure*}
\subsection{Motivation}

Compared with the equirectangular map, the cubemap contains far less distortion, significantly decreasing depth estimation difficulty. However, when converting to an equirectangular map, estimating each face of the cubemap can easily lead to depth discontinuities and an inability to recover missing pixels accurately, as shown in Fig.2.


To solve this problem, another reference depth map is used to smooth the discontinuous depth. The original cube and the $\phi$ rotated one completely revise the discontinuity because they have complementary boundary discontinuity areas and a consistent context (illustrated in Fig. 1, center). Besides, we observe that when $\phi$ is equal to 45°, the revision effect comes to the best. 

We define a rotation operation $\boldsymbol{R}$ representing the corresponding horizontal rotation. Besides, equirectangular-to-cube and cube-to-equirectangular transformations are donated as $\mathfrak{T}$ and $\mathfrak{T}^{-1}$, respectively. Given an equirectangular representation input $I_{equi}\in  \mathbb{R}^{W\times H \times3}$, the whole process can be expressed as follows:
\begin{eqnarray}
D_{equi} &=& \mathfrak{T}^{-1}(\boldsymbol{f}_{DCDE}(\mathfrak{T}(I_{equi}))), \\
D_{equi,\phi} &=& \boldsymbol{R}^{-1}(\mathfrak{T}^{-1}(\boldsymbol{f}_{DCDE}(\mathfrak{T}(\boldsymbol{R}(I_{equi},\phi))))),
\end{eqnarray}
where $D_{equi},D_{equi,\phi}$ represents the output of DCDE module.

That is feasible but results in a substantial computational overhead. To optimize the algorithm, we quickly implement the rotation-based process by moving the map blocks (illustrated in Fig. 4) according to the conversion relationship between the sample points and spherical coordinates.  
\subsection{Dual-Cubemap Depth Estimation Module}
The dual-cubemap depth module is established using two independent and parallel branches, and each branch shares the same architecture. In each branch, we adopt the pre-trained ResNet-50~\cite{2016Deep} to extract features and up-projection~\cite{2020BiFuse} to generate the depth from features.
\begin{figure}[!t]
  \centering
  \includegraphics[width=.35\textwidth]{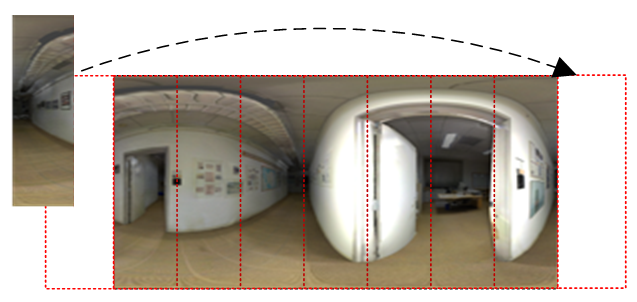} 
  \caption{If $\phi$ = 45°, each rotation is equivalent to a circular recursive shift of one-eighth block of the image.} 
  \label{fig_cut} 
\end{figure}
However, the discontinuous depth will be caused due to the discontinuous features in each cube face. To eliminate depth discontinuity at the feature-level, we design a boundary-aware block to promote the interaction between the two branches. Specifically, we convert the cube feature maps that output from layer $i$ into equirectangular format to expose the discontinuity and fuse the two branches' features. The input of layer $i+1$ of the two branches can be respectively expressed as:
\begin{eqnarray}
f_{1}^{i+1}&=&f_{1}^{i}+ \mathfrak{T}(\boldsymbol{R}^{-1}(\mathfrak{T}^{-1}(f_{2}^{i}))\otimes \mathfrak{T}^{-1}(f_{1}^{i})),\\
f_{2}^{i+1}&=&f_{2}^{i}+ \mathfrak{T}((\mathfrak{T}^{-1}(f_{2}^{i})\otimes \boldsymbol{R}(\mathfrak{T}^{-1}(f_{1}^{i}))),
\end{eqnarray}
where subscript 1, 2 represent the first and second branch, respectively, and $\otimes$ is Hadamard product. We get two estimated depth maps in cubemap representation from this module.
\subsection{Boundary Revision Module}
The depth estimated from the previous module is merely a coarse result with obvious boundary and depth discontinuity, another BR module is designed to revise the boundary at pixel-level further. We adopt an encoder-decoder architecture to implement this module. The network contains 10 convolution blocks, as shown in Fig. 3(right), and each of which comprises two different convolution layers. A maxpooling or deconvolution is adopted every two layers. Besides, skip connections are applied to connect the low-level and high-level features with the same resolution by a 3×3 convolutional layer to prevent the gradient vanishing problem and information imbalance in training~\cite{Zhou2018UNet}. 

\subsection{Objective function}

Depth mutations often appear at the boundary of the object. Therefore, to make the depth estimation at that area more accurate, we introduce gradient loss~\cite{DBLP:journals/corr/abs-1810-01849}. Besides, the reverse Huber~\cite{Cavazza2016Active} or Berhu loss~\cite{Laina2016Deeper} is also added in our objective function:
\begin{eqnarray}
\beta  = \begin{cases}
\left | x \right |\qquad  \left | x \right |\leq c,\cr
\frac{x^{2}+c^{2}}{2c} \qquad \left | x \right |>  c.\cr
\end{cases}
\end{eqnarray}
\begin{eqnarray}
L_{grad}(y,\hat{y}) = \frac{1}{n}\sum_{p}^{n}\left | g_{x}(y_{p},\hat{y_{p}}) \right |+\left | g_{y}(y_{p},\hat{y_{p}}) \right |
\end{eqnarray}
And our loss function can be written as:
\begin{eqnarray}
L_{\beta 1}& = & \sum_{i\epsilon p}\beta(\mathfrak{T}^{-1}(D_{1}^{i}),D_{gt}^{i}), 
\end{eqnarray}
where $D_{1}$ is the prediction produced by the first branch; $D_{gt}$ is the ground truth; and P represents all the valid value in the ground truth map.
\begin{eqnarray}
L_{\beta 2}& = &\sum_{i\epsilon p}\beta(\mathfrak{T}^{-1}(D_{2}^{i}),D_{gt}^{i}),
\end{eqnarray}
where $D_{2}$ is the prediction produced by the second branch.
\begin{eqnarray}
L_{\beta f}& = &\sum_{i\epsilon p}\beta(D_{f}^{i},D_{gt}^{i}),
\end{eqnarray}where $D_{f}$ is the final prediction.
\begin{table*}[!ht]
\begin{center}
\caption{Quantitative comparison on Matterport3D and Stanford2D3D Datasets.} \label{tab:cap1}
\begin{tabular}{|c|c|c|c|c|c|c|c|}
  \hline
  Dataset & Method & $MAE\downarrow$&$RMSE\downarrow$&$RMSE(log)\downarrow$&$\delta_{1}\uparrow$&$\$delta_{2}\uparrow$&$\delta_{3}\uparrow$
  \\
  \hline
   & FRCN~\cite{Laina2016Deeper} & 0.4008&0.6704&0.1244&0.7703 &0.9174&0.9617\\
   \cline{2-8}
  & OmniDepth~\cite{Zioulis2018OmniDepth} &0.4838&0.7643&0.1450&0.6830&0.8794&0.9429 \\
   \cline{2-8}
    Matterport3D&BiFuse(cp)~\cite{Cheng2018Cube} &0.3929&0.6628&-&-&-&-\\
    \cline{2-8}
 &BiFuse(sp)~\cite{2020BiFuse}&\textbf{{\color{blue}0.3470}}&\textbf{{\color{blue}0.6295}}&\textbf{{\color{blue}0.1281}}&\textbf{{\color{blue}0.8452}}&\textbf{{\color{blue}0.9319}}&\textbf{{\color{blue}0.9632}}\\
  \cline{2-8}
    & Ours &\textbf{{\color{red}0.2552}}&\textbf{{\color{red}0.5381}}&\textbf{{\color{red}0.1075}}&\textbf{{\color{red}0.8896}}&\textbf{{\color{red}0.9722}}&\textbf{{\color{red}0.9893}}\\
  \hline
   & FRCN~\cite{Laina2016Deeper}&0.3428&0.5774&0.1100&0.7230&0.9207&0.9731\\
    \cline{2-8}
 & OmniDepth~\cite{Zioulis2018OmniDepth} &0.3743&0.6152&0.1212&0.6877&0.8891&0.9578 \\
  \cline{2-8}
   Stanford2D3D&BiFuse(cp)~\cite{Cheng2018Cube} &0.2588&0.4407&-&-&-&-\\
   \cline{2-8}
&BiFuse(sp)~\cite{2020BiFuse} &\textbf{{\color{blue}0.2343}}&\textbf{{\color{red}0.4142}}&\textbf{{\color{red}0.0787}}&\textbf{{\color{blue}0.8660}}&\textbf{{\color{blue}0.9580}}&\textbf{{\color{blue}0.9860}}\\
 \cline{2-8}
    & Ours &\textbf{{\color{red}0.1876}}&\textbf{{\color{blue}0.4573}}&\textbf{{\color{blue}0.0894}}&\textbf{{\color{red}0.9039}}&\textbf{{\color{red}0.9809}}&\textbf{{\color{red}0.9886}}\\
  \hline
\end{tabular}
\end{center}
\end{table*}
 \begin{figure*}
  \centering
  \includegraphics[width=\textwidth]{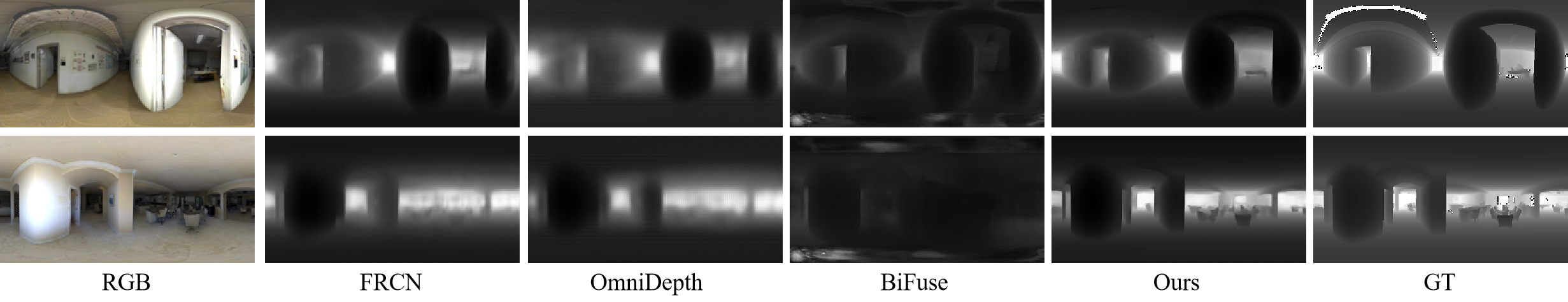} 
  \caption{Qualitative comparison on Stanford2D3D (Up) and Matterport3D (Down).} 
  \label{fig_cut} 
\end{figure*}
\begin{eqnarray}
L_{grad}(D_{gt},D_{f}) = \frac{1}{n}\sum_{i \in p}^{n}\left | g_{x}(D_{gt}^{i},D_{f}^{i}) \right |+\left | g_{y}(D_{gt}^{i},D_{f}^{i}) \right |,
\end{eqnarray}where $g_{x}$, $g_{y}$ represent horizontal gradient and vertical gradient, respectively; and n is the total number of valid values from the ground truth.
The total objective function is:
\begin{eqnarray}
L_{total} = \lambda _{\beta 1}L_{\beta 1}+\lambda _{\beta 2}L_{\beta 2}+\lambda _{\beta f}L_{\beta f}+L_{grad},
\end{eqnarray}
where $\lambda _{\beta 1},\lambda _{\beta 2},\lambda _{\beta f}$ are the weight of each part. While training our model,  the weights are set to $\lambda _{\beta 1} = 0.1,\lambda _{\beta 2} = 0.1,$ and $\lambda _{\beta f} = 0.8$.

\section{Experiment}

To evaluate our framework, extensive experiments are conducted to compare the proposed method with other state-of-the-art approaches and validate the effectiveness of each module.
\subsection{Basic Settings}

We conduct our experiments on an Nvidia RTX 2080 MQ GPU. Adam optimizer is chosen, and the batch size is set as 4. The learning rate is $5\times10^{-4}$ for the first 20 epochs and then adjusted to $1\times10^{-4}$ for the next 20 epochs. In addition, a model dynamic storage mechanism is set up to pick the best performance model when the training is stable.

\subsection{Metrics and Datasets}

We adopt the popular evaluation metrics used by previous works~\cite{cheng2020odecnn,Zioulis2018OmniDepth} for performance analysis, including MAE, RMSE, RMSE(log),and $\delta_{t}$, $t\in(1.25,1.25^{2},1.25^{3})$. 

The experiments are carried out on Matterport3D and Stanford2D3D, two real indoor datasets. The depth values of the two datasets are collected by 3D cameras or sensors, which means there is noise or missing values in certain areas. We follow the official guidance to filter the unavailable images and limit the data range to filter out invalid values. This process can avoid the noise interfering in the experiment (According to the official description, the depth collection values range in 0 m to 10 m). Finally, we get 6319 images for training, 865 for validating, and 1632 for testing from Matterport3D; and 898 for training, 81 for validating, 365 for testing from Stanford2D3D.
\begin{table}[t]
\begin{center}
\caption{Ablation comparison on Stanford2D3D} \label{tab:cap1}
\begin{tabular}{|c|c|c|c|c|c|c|c|}
  \hline
  $I_{cube1}$&$I_{cube2}$&BR&GL&$MAE\downarrow$&$\delta_{1}\uparrow$
  \\
  \hline
  \checkmark&\ding{55} &\ding{55} & \ding{55}&0.5091&0.5264 \\
  \checkmark&\checkmark&\ding{55} & \ding{55}&0.2414&0.8519\\
   \checkmark&\checkmark&\checkmark& \ding{55}&0.2103&0.8894\\
\checkmark&\checkmark&\checkmark&\checkmark&\textbf{0.1876}&\textbf{0.9039}\\
\hline
\end{tabular}
\end{center}
\end{table}
\subsection{Quantitative and Qualitative Analysis}
We exhibit quantitative comparisons on the two datasets in Table 1, where we mark the 1st best solution in red and the 2nd best solution in blue. The results show that our method outperforms all the other methods in every metric on Matterport3D. On Stanford2D3D, we perform 1st in MAE, $\delta_{1}$, $\delta_{2}$, $\delta_{3}$ and 2nd in the remaining two metrics. Since there is more noise in Stanford2D3D, MAE can better measure the performance of the model than RMSE. Overall, our predictions are on average 4$\%$ more accurate in $\delta_{1}$ than that reported by BiFuse~\cite{2020BiFuse}. The comparisons experimental validates the effectiveness of our dual-cubemap approach. The two cubemaps not only focus on the distortion-free regions but also retain the information from the large FoV of the panorama as much as possible, which both contribute to improving estimation accuracy.

Fig. 5 demonstrates the qualitative comparisons of the two datasets. Note that since BiFuse~\cite{2020BiFuse} does not release the training code, we can only use their model trained on 360D~\cite{Zioulis2018OmniDepth} for qualitative analysis. In general, our results are visually more clear and sharper around boundaries, which can be attributed to the complementary information from the dual-cubemap and gradient loss constraint on edges.
\subsection{Ablation Experiment}
In this section, ablation studies are performed on the following components:
1) $I_{cube1}$: the 0° rotated cubemap branch; 2) $I_{cube2}$: the 45° rotated cubemap branch; 3) BR: boundary revision module; 4) GL: gradient loss. Especially, 1) and 2) are chosen simultaneously to represent the dual-cubemap depth module.

From Table 2, we can observe that one branch with a cubemap representation image as input performs much worse than others because a single cubemap has a smaller FoV than an equirectangular map. Furthermore, the supplement information from the other cubemap has dramatically improved the performance. Since GL is only used to sharpen edges~\cite{DBLP:journals/corr/abs-1810-01849} without eliminating the boundary discontinuity, the BR module is proposed. It finally revises the discontinuity of boundaries and outperforms 13$\%$ in terms of MAE. 

As shown in Fig. 6, the depth estimated from our dual-cubemap is much better than that from a single cubemap, but there is still visible discontinuity. Moreover, the BR module revises the discontinuity effectively, and gradient loss makes the results more accurate on edges.
\begin{figure}[!t]
  \centering
  \includegraphics[width=.35\textwidth]{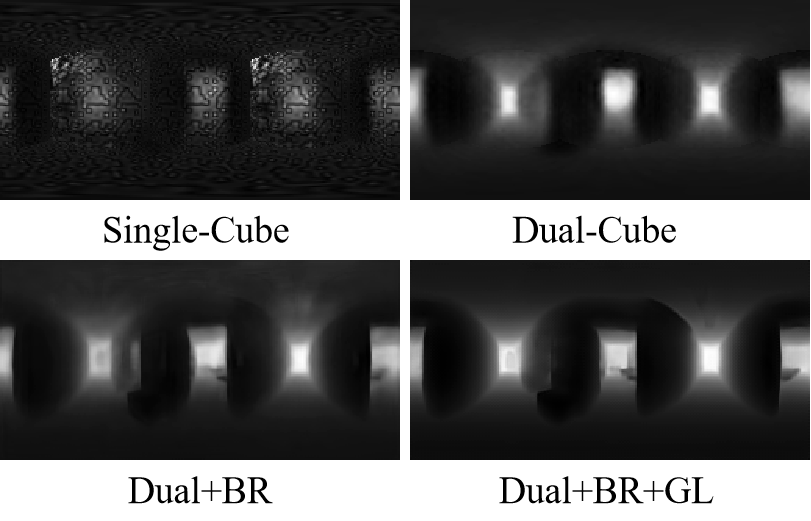} 
  \caption{Qualitative ablation comparison on Stanford2D3D.} 
  \label{fig_cover} 
\end{figure}
\section{Conclusion}
In this paper, we propose an end-to-end trainable omnidirectional depth estimation network consisting of two modules for depth estimation and boundary revision, respectively. Since the cubemap representation reduces the influence of distortion, we entirely rely on this representation for depth estimation. To extend the FoV of the cubemap presentation and revise the boundary, we adopt a rotation-based two-cubemap strategy. We reconstruct the large FoV depth map using the two cubemaps with consistent context but different discontinuous boundary regions. Experimental results demonstrate that our approach outperforms the current state-of-the-art methods.
\\
\textbf{Acknowledgement.} This work was supported by National Natural Science Foundation of China (No.61772066, No.61972028), and the Fundamental
Research Funds for the Central Universities (No.2018JBZ001).


\bibliographystyle{IEEEbib}
\bibliography{icme2021template}

\end{document}